\documentclass[10pt,twocolumn,letterpaper]{article}

\usepackage[pagenumbers]{cvpr}      

\usepackage[dvipsnames]{xcolor}

\definecolor{brickred}{rgb}{0.8, 0.25, 0.33}
\definecolor{yaleblue}{rgb}{0.06, 0.3, 0.57}
\definecolor{vividauburn}{rgb}{0.58, 0.15, 0.14}

\definecolor{cvprblue}{rgb}{0.21,0.49,0.74}
\definecolor{mygreen}{RGB}{129, 169, 102}
\definecolor{mypink}{RGB}{215, 189, 189}
\definecolor{myyellow}{RGB}{225, 204, 105}

\usepackage{multirow}
\usepackage{pifont}
\usepackage[pagebackref,breaklinks,colorlinks,citecolor=cvprblue]{hyperref}
\newcommand\mypar[1]{\par\vspace{1.0mm}\noindent\textbf{#1}\;\;}

\expandafter\def\expandafter\normalsize\expandafter{%
    \normalsize%
    \setlength\abovedisplayskip{3pt}%
    \setlength\belowdisplayskip{3pt}%
    \setlength\abovedisplayshortskip{-3pt}%
    \setlength\belowdisplayshortskip{3pt}%
}

\newcommand\mem{\mathbf{M}}
\newcommand\attn{\text{Attn}}
\newcommand\attnQ{\text{Q}}
\newcommand\attnK{\text{K}}
\newcommand\attnV{\text{V}}
\newcommand\refeq[1]{Eqn.~\ref{#1}}
\newcommand\reftab[1]{Table~\ref{#1}}
\newcommand\reffig[1]{Fig.~\ref{#1}}
\newcommand\refsec[1]{Sec.~\ref{#1}}
\newcommand\jaccd{\mathcal{J}}
\newcommand\jaccdLast{\jaccd_{\mathrm{tr}}}
\newcommand\contour{\mathcal{F}}
\newcommand{\cmark}{\ding{51}}%
\newcommand{\secspace}{-2.0mm}%
\def\paperID{6517} 
\def\confName{CVPR}
\def\confYear{2024}

\def\ourmodel{RMem\xspace}

\title{RMem: Restricted Memory Banks Improve Video Object Segmentation\vspace{-6mm}}

\author{Junbao Zhou$^{*}$ \quad Ziqi Pang$^*$ \quad Yu-Xiong Wang \\
University of Illinois Urbana-Champaign \\
{\tt\small \{junbaoz,ziqip2,yxw\}@illinois.edu}
}

\begin{document}
\maketitle

\begin{NoHyper}
\def\thefootnote{*}\footnotetext{Equal contribution.}
\end{NoHyper}
\def\thefootnote{\arabic{footnote}}

\begin{abstract}
\vspace{-2mm}
With recent video object segmentation (VOS) benchmarks evolving to challenging scenarios, we revisit a simple but overlooked strategy: restricting the size of memory banks. This diverges from the prevalent practice of expanding memory banks to accommodate extensive historical information. Our specially designed ``memory deciphering'' study offers a pivotal insight underpinning such a strategy: expanding memory banks, while seemingly beneficial, actually increases the difficulty for VOS modules to decode relevant features due to the confusion from redundant information. By restricting memory banks to a limited number of essential frames, we achieve a notable improvement in VOS accuracy. This process balances the importance and freshness of frames to maintain an informative memory bank within a bounded capacity. Additionally, restricted memory banks reduce the training-inference discrepancy in memory lengths compared with continuous expansion. This fosters new opportunities in temporal reasoning and enables us to introduce the previously overlooked ``temporal positional embedding.'' Finally, our insights are embodied in ``\ourmodel'' (``R'' for restricted), a simple yet effective VOS modification that excels at challenging VOS scenarios and establishes new state of the art for object state changes (on the VOST dataset) and long videos (on the Long Videos dataset). Our code and demo are available at \href{https://restricted-memory.github.io/}{https://restricted-memory.github.io/}.
\end{abstract}
\vspace{-4mm}
\section{Introduction}
\label{sec:intro}
\vspace{\secspace}

The rapid progress of video object segmentation (VOS) algorithms has motivated the creation of more challenging benchmarks, as exemplified by VOST~\cite{tokmakov2023breaking} on more complicated videos with significant \emph{object state changes} and the Long Videos dataset~\cite{liang2020video} featuring extremely \emph{long} duration. These benchmarks elevate the spatio-temporal modeling and prompt us to reassess conventional VOS designs: \emph{can learning-based VOS modules effectively decipher historical information in such challenging scenarios}? 

\begin{figure}
    \centering
    \includegraphics[width=0.88\linewidth]{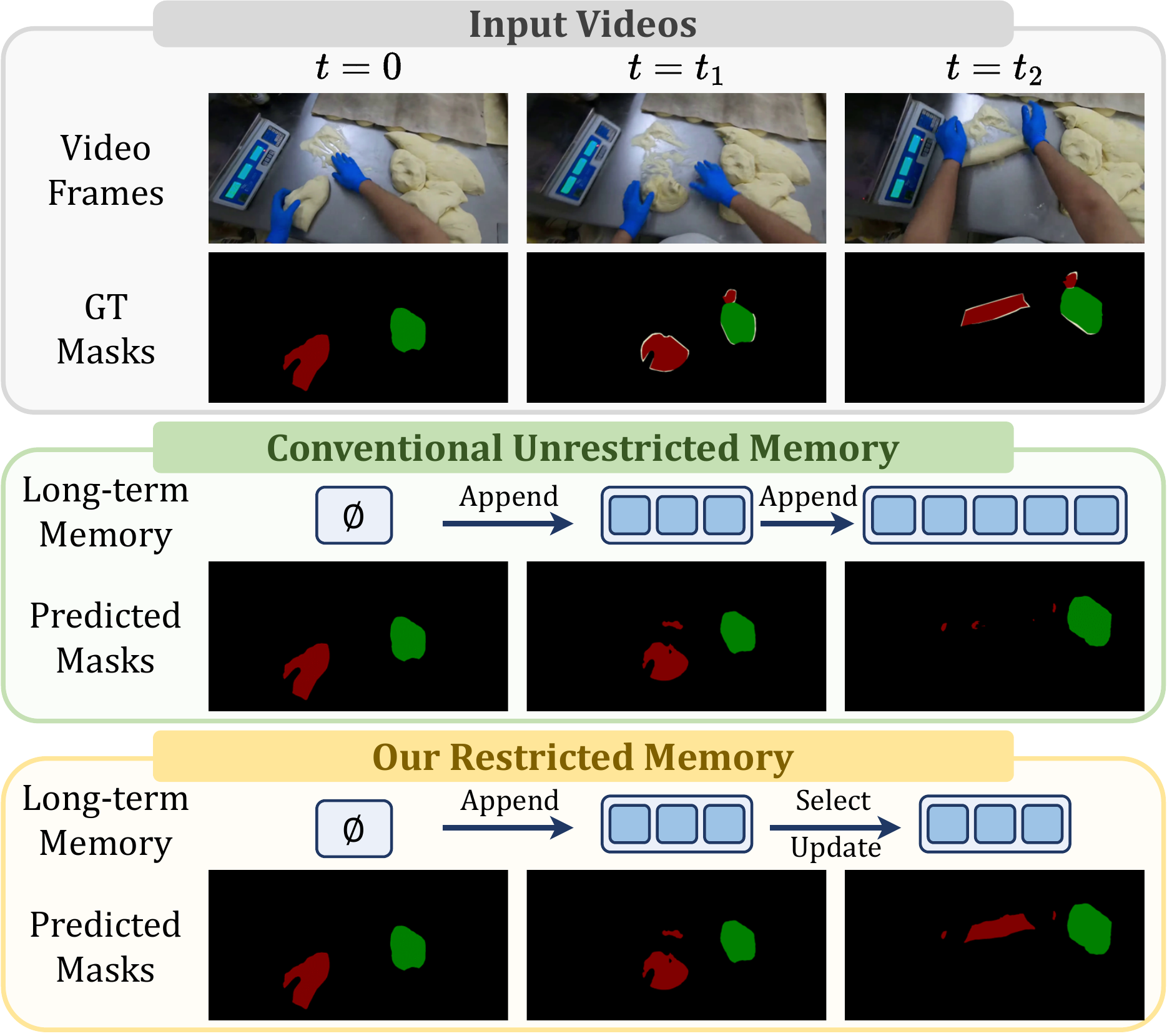}
    \vspace{-3mm}
    \caption{In light of challenging object state changes~\cite{tokmakov2023breaking, xue2024learning, yu2023video}, we rethink the conventional VOS approach of continuously accumulating the features into memory banks: despite capturing all the information, it complicates the deciphering of relevant features. Conversely, restricted memory banks significantly enhance VOS.}
    \vspace{-6mm}
    \label{fig:teaser}
\end{figure}

To delve into this issue, it is essential to focus on \emph{memory banks}, which are central to storing past features and feeding input to VOS modules, and are fundamental in the memory-based VOS framework~\cite{cheng2022xmem, yang2021aot, cheng2021rethinking}. Typically, the memory banks are managed via the simple intuition of \emph{expansion}, continuously \emph{appending} newly sampled frames as the video progresses. While this approach is intended to encompass all historical information, thereby enhancing VOS, we realize its potential limitation: as videos become longer or more complex, these expanding memory banks may overwhelm the capability of VOS modules to discern reliable features.

We investigate this hypothesis by conducting a pilot study, named ``\emph{memory deciphering},'' to quantify the decoding capability of VOS modules. In our analysis, we continue to use object segmentation as the proxy to VOS, but shift the prediction target to decoding \emph{the object mask at the initial frame (frame 0)} from the memory bank. This choice is deliberate based on the principle of controlling variables: (1) In the VOS framework, the information of frame 0 is implicitly propagated to subsequent frames, ensuring the presence of relevant information for decoding; (2) This prediction target is consistent across frames and allows for a fair comparison of decoding efficacy under varying memory sizes. Intuitively, the later frames have \emph{rigorously richer} information than the earlier frames because of a larger memory bank, and are thus expected to produce better decoding results. However, our observation shows the opposite: \emph{the effectiveness of VOS modules in deciphering information diminishes with increasingly large memory banks}. Intriguingly, this degradation can be mitigated by selecting a small number of relevant frames in the memory bank, and we observe a significantly better concentration of attention scores on relevant frames and regions. Therefore, our systematic study reveals a pivotal insight: \emph{the expansion of memory banks complicates the deciphering of VOS modules} primarily due to \emph{redundant information}.

Inspired by such an insight, we validate its practical significance through a simple approach: \emph{restricting memory banks to a fixed number of frames}. Our concise memory bank facilitates better spatio-temporal modeling and adaptation to object transformation according to the analysis of complex object state changes~\cite{tokmakov2023breaking}, as illustrated in Fig.~\ref{fig:teaser}. The effectiveness of our method stems from a curated memory concisely focusing the attention of VOS modules on relevant information. Based on this, we delve into the updating process when new features arrive. Our strategy balances the relevance and freshness of frame features, drawing inspiration from the upper confidence bound (UCB) algorithm~\cite{auer2002using} from multi-arm bandit problems.

In addition to enhancing the accuracy, restricted memory banks reduce \emph{ discrepancies in memory lengths} between training and inference when compared with conventional methods. Typically, VOS modules are trained on short clips with a few memory frames, so our restricted memory bank better aligns with this setup, even when handling significantly longer videos during inference. This alignment opens up opportunities to revisit techniques relying on temporal synchronization between training and inference. As a compelling example, we introduce \emph{temporal positional embedding} to explicitly capture the ordering of memory features -- a critical aspect often overlooked by previous methods -- leading to superior temporal reasoning.

In conclusion, we make the following contributions:
\begin{enumerate}
    \item We introduce the novel \emph{memory deciphering} analysis to systematically reveal the drawbacks of expanding memory banks for VOS modules in decoding information.
    \item Our revisit of \emph{restricting memory banks} notably enhances VOS accuracy for challenging cases, cooperated with a memory update strategy balancing the relevance and freshness of frames.
    \item Benefiting from smaller training-inference gaps, we introduce the previously overlooked \emph{temporal positional embedding} to capture the order of memory frames.
\end{enumerate}

\noindent Collectively, our insights lead to a simple yet strong VOS method: ``\ourmodel,'' which is \emph{plug-and-play} for memory-based VOS methods. Our extensive experiments show its strengths and establish new state of the art on VOST~\cite{tokmakov2023breaking} for object state changes and the Long Videos dataset~\cite{liang2020video}.
\section{Related Work}
\vspace{-3mm}

\mypar{VOS benchmarks.}  VOS has evolved through several benchmarks. DAVIS~\cite{perazzi2016benchmark, pont20172017} is the first exhibiting diversity and quality, surpassing early benchmarks~\cite{brox2010object, li2013video, tsai2012motion}. YoutubeVOS~\cite{xu2018youtube} further scales up by collecting more videos. Although they have enabled great progress in VOS, their limited difficulty and video lengths have spurred more challenging datasets. For example, the average duration in LVOS~\cite{hong2023lvos} is more than 500 frames and the Long Videos dataset~\cite{liang2020video} further extends it to over 1,000 frames, and MOSE~\cite{ding2023mose} increases the difficulty by selecting videos with crowds and occlusions. To evaluate our insight on the most demanding scenarios, we highlight \emph{object state changes} involving noticeable transformations in the existence, appearance, and shapes. Studies on state changes, \emph{e.g.}, VSCOS~\cite{yu2023video}, mostly utilize ego-centric datasets~\cite{damen2018scaling, visor2022, grauman2022ego4d}. In this paper, we primarily select the recent VOST~\cite{tokmakov2023breaking}. It combines multiple datasets and provides accurate annotations. Notably, VOST shows higher complexity and longer duration than previous YoutubeVOS and DAVIS. We mainly concentrate on the challenging benchmarks.

\mypar{Memory-based VOS.} Memory banks are fundamental for VOS. Earlier approaches~\cite{caelles2017one, voigtlaender2017online, bhat2020learning, maninis2018video, robinson2020learning} treat VOS as online learning and finetune networks with memorized features. Some others~\cite{chen2018blazingly, hu2018videomatch, yang2018efficient, yang2020collaborative, yang2021collaborative, voigtlaender2019feelvos} approach VOS as template matching but struggle with occluded or dynamically changing objects. Consequently, recent methods mostly focus on memory reading via either pixel-level or object-level attention~\cite{vaswani2017attention}. Object-level memory reading~\cite{athar2022hodor, athar2023tarvis, cheng2023putting}, inspired by Mask2Former~\cite{cheng2022masked}, excels at efficiency. However, it is less effective for delicate masks or complex scenarios, \emph{e.g.}, VOST~\cite{tokmakov2023breaking}, where the objects are frequently small or cluttered. In comparison, pixel-wise memory reading~\cite{oh2018fast, seong2020kernelized, duke2021sstvos, liang2020video, cheng2022xmem, xie2021efficient, cheng2021rethinking, yang2021aot, yang2022decoupling} is more adopted for its reliable segmentation and it typically associates the current frame to memory features with attention. Our work differs from previous studies by focusing more on the general insights of \emph{drawbacks of expanding memory banks} and plug-and-play strategies to mitigate such issues, instead of dedicated memory reading architectures. 

\begin{figure*}[ht]
    \centering
    \includegraphics[width=0.95\textwidth]{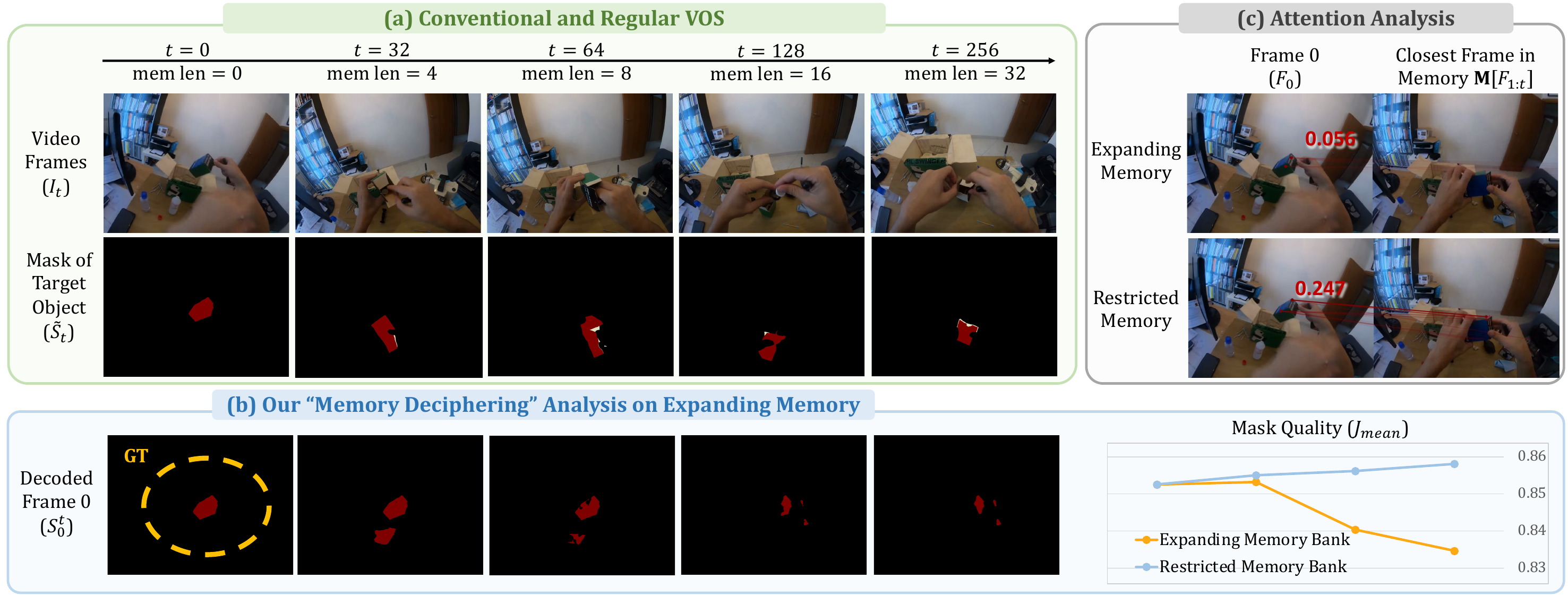}
    \vspace{-3mm}
    \caption{Sketch of Pilot Study. Our \emph{memory deciphering} analysis emulates \emph{decoding the mask on frame 0 from the memory bank features} to quantify the impact of a growing memory bank on VOS modules, where the ``desired results'' in the figure are the ground truth. For a video shown in Block~\textbf{(a)}, we visualize its decoding results in Block~\textbf{(b)}: the masks degrade both quantitatively (yellow curve) and qualitatively, deviating from the desired results. However, selecting a set of concise frames mitigates this issue (blue curve in Block~\textbf{(b)}). Therefore, we conjecture that the drawback of a growing memory bank lies in confusing the attention of VOS modules. In Block~\textbf{(c)}, we use red lines to indicate highly weighted associations in attention, with thickness denoting the attention score values. As illustrated, the query $F_0$ focuses less on its most relevant frame after the memory bank expands, with the attention score dropping from 0.247 to 0.056. (2$^{\text{nd}}$ row shows ground-truth masks $\widetilde{S}_t$ as the reference. $\mathcal{J}_{\mathrm{mean}}$ is the average Jaccard between $S_0^t$ and $\widetilde{S}_0$ over all videos.)}
    \vspace{-5mm}
    \label{fig:pilot_experiments}
\end{figure*}

\mypar{Restricted Memory Banks.} Previous studies approach restricting memory banks mostly from the efficiency aspect~\cite{liang2020video, li2020fast, cheng2022xmem}. A notable representative, XMem~\cite{cheng2022xmem}, adopts a hierarchical architecture with customized modifications like memory potentiation. In contrast to prior efforts, our work explicitly reveal and highlight the \emph{accuracy} benefits of restricted memory banks through reducing redundant information, rather than emphasizing \emph{efficiency}. Moreover, our \ourmodel demonstrates such an insight with a \emph{simple plug-and-play} enhancement to the VOS framework, avoiding any noticeable increase or reliance on special operators as in XMem. We further suggest that RMem's benefits are not constrained to VOS, where recent works~\cite{qian2024streaming} applying large language models to long videos notice the similar benefits of condensing memory and selecting frames.
\vspace{\secspace}
\section{Pilot Study: Memory Deciphering Analysis}
\label{sec:pilot}
\vspace{-3mm}

This section devises our pilot experiments on how an expanding memory bank influences the decoding capability of VOS modules. Our design emulates the task of VOS but makes several modifications guided by the principle of controlling variables: the prediction targets and VOS modules are aligned across our pilot experiments, while only the frames in the memory bank vary. Such a comparison enables a clean analysis and reveals the core insight: VOS modules have limited capability to decode a growing memory bank. 

\mypar{Notation and Formulation of VOS.}
We consider the existing VOS framework as a memory-based encoder-decoder network: the encoder $\mathbf{E}(\cdot)$ is a visual backbone encoding the image $I_t$ at frame $t$ into the feature $F_t$; and then, the decoder $\mathbf{D}(\cdot)$ converts $F_t$ into the segmentation $S_t$ via reading the features stored in the memory $\mathbf{M}[F_{0:t-1}]$, as below,
\begin{equation}
    \label{eq:encoder_decoder}
    F_t = \mathbf{E}(I_t),\quad S_t = \mathbf{D}(F_t, \mathbf{M}[F_{0:t-1}]).
\end{equation}
Here, $\mathbf{M}[F_{0:t-1}]$ generally comes from saving the features at a certain frequency~\cite{yang2021aot, cheng2021rethinking, li2020fast}, and the VOS decoder is usually special transformers~\cite{vaswani2017attention}, \emph{e.g.}, LSTT in AOT~\cite{yang2021aot}. The final objective of VOS is to minimize the difference between the predicted mask $S_t$ and ground truth $\widetilde{S}_t$.

\begin{figure*}
    \centering
    \includegraphics[width=0.82\textwidth]{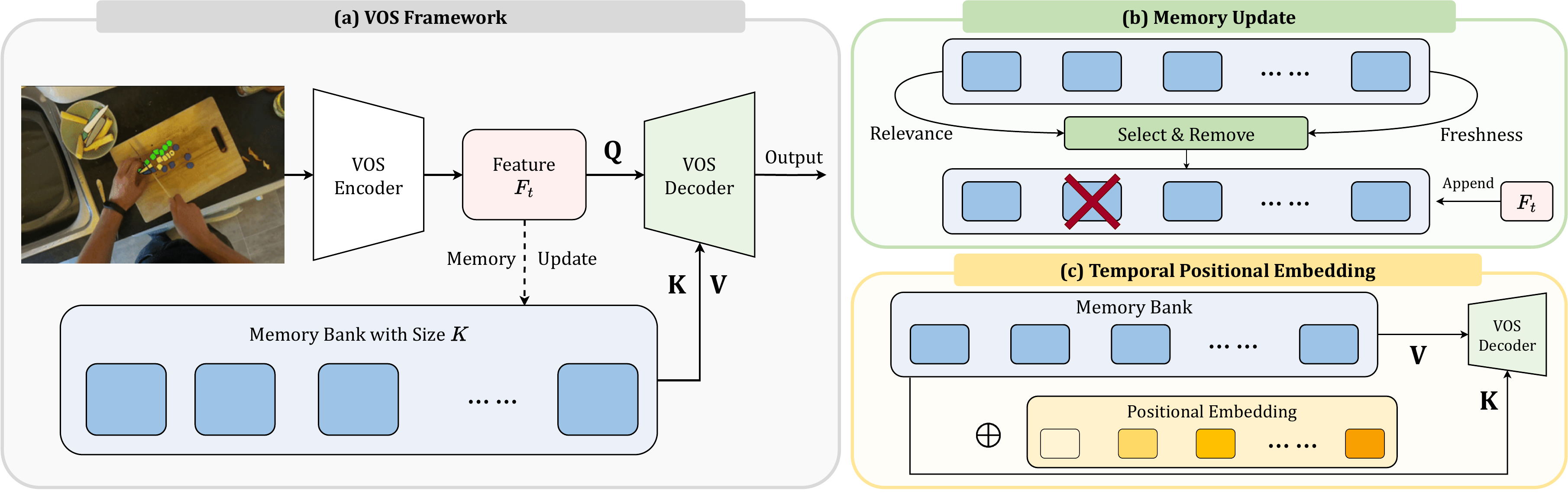}
    \vspace{-3mm}
    \caption{\ourmodel Overview. \textbf{(a)} \ourmodel revisits \emph{restricting memory banks} to enhance VOS (Sec.~\ref{sec:restrict_memory}), motivated by the insight from our pilot study. \textbf{(b)} To maintain an informative memory bank, we balance both the relevance and freshness of frames when updating the latest features (Sec.~\ref{sec:mem_update}). \textbf{(c)} Benefiting from smaller memory size gaps between training and inference, we introduce previously overlooked temporal positional embedding to encode the orders of frames explicitly (Sec.~\ref{sec:mem_temporal}), which enhances spatio-temporal reasoning. }
    \vspace{-5mm}
    \label{fig:method}
\end{figure*}

\mypar{Design of Our Memory Deciphering Analysis.} Our pilot study separates the variables of the VOS module $\mathbf{D}(\cdot)$ and the prediction target $\widetilde{S}_t$ to clearly analyze the influence of the memory bank $\mathbf{M}[F_{0:t-1}]$ under a controlling variable setting. Therefore, we purposefully design our memory deciphering analysis as \emph{decoding the mask of the initial frame (frame 0) from the features stored in the memory bank}. 

More precisely, our pilot study is formulated as,
\begin{equation}
    \label{eq:memorization}
    S_0^t = \mathbf{D}'(F_0, \mathbf{M}[F_{1:t}]),
\end{equation}
where $\mathbf{D}'(\cdot)$ is an additional VOS decoder trained for the objective in Eqn.~\ref{eq:memorization}. In practice, we use the original VOS decoder $\mathbf{D}(\cdot)$ to conduct regular VOS as Eqn.~\ref{eq:encoder_decoder}, and then employ $\mathbf{D}'(\cdot)$ only for deciphering the mask $S_t^0$ for frame 0, to avoid influencing the original VOS. $\mathbf{M}[F_{1:t}]$ contains the stored features between frames 1 to $t$. Note that the feature of frame 0 is excluded from the input $\mathbf{M}[F_{1:t}]$ to avoid $\mathbf{D}'$ from trivially relying on single-frame memory.

Before delving into the experiments, we emphasize our reasons for choosing this formulation. (1) \emph{Presence of relevant information}. The procedure in Eqn.~\ref{eq:encoder_decoder} resembles propagating the masks from historical frames to the current frame $t$, indicating that $\mathbf{M}[F_{1:t}]$ contains the information about the mask at frame 0. Therefore, decoding the mask on frame 0 from $\mathbf{M}[F_{1:t}]$ is not a random guess, but should achieve high-quality results. (2) \emph{Identical prediction target}. Our prediction target remains identical for every frame and varying memory size. (3) \emph{Cooperating with regular VOS}. We utilize $\mathbf{D}'(\cdot)$ as a stand-alone VOS decoder so that the original VOS process remains unchanged and our pilot study can utilize the same memory bank. 

\mypar{Implementation.} We select the recent VOST~\cite{tokmakov2023breaking} dataset to highlight challenging \emph{object state changes}. Its long video duration and complex scenarios push the limits of VOS decoders in deciphering memory. Then we adopt AOT~\cite{yang2021aot} as the VOS encoder-decoder, a popular baseline and the top method on VOST. Emulating Eqn.~\ref{eq:memorization}, we initialize $\mathbf{D}'(\cdot)$ from AOT's pretrained decoder $\mathbf{D}(\cdot)$, and then supervise $S_t^0$ with a segmentation loss between the ground truth $\widetilde{S}_0$. More implementation details are in \refsec{sec:supp_implementation_details}.

\mypar{Hypothesis and Expectations.} With an expanding memory bank, the information in $\mathbf{M}[F_{1:t}]$ becomes \emph{rigorously richer} at later frames while the prediction target is unchanged. Therefore, we naturally expect the decoded mask $S^t_0$ to illustrate stable or better accuracy at later frames, assuming that the VOS decoder $\mathbf{D}(\cdot)$ is capable of extracting the relevant features from an increasingly large $\mathbf{M}[F_{1:t}]$.

\mypar{Results and Analysis.} Contrasting the expectation above, we observe that masks $S_0^t$ degrade with a growing memory bank, as shown in Fig.~\ref{fig:pilot_experiments}~(b). To verify that the growing memory bank is indeed the cause of degradation, we empirically bound the memory bank to 8 frames containing the most relevant and latest information, intuitively: first 7 frames and the latest frame in $\mathbf{M}[F_{1:t}]$. According to the blue curve in Fig.~\ref{fig:pilot_experiments}~(b), \emph{restricting the memory only to store concise features} effectively avoids degradation.

Inspired by addressing the degradation issue, we propose that the \emph{redundant information} is the main negative impact of an expanding memory bank. Otherwise, the degradation should not disappear simply after we select a subset of intuitively relevant frames. More specifically, this closely relates to how VOS methods utilize attention mechanisms to read from memory banks, where the redundant features decrease the attention scores on relevant frames. As direct evidence, we analyze the attention scores for decoding $S_0^t$ in Fig.~\ref{fig:pilot_experiments}~(c) and observe that the attention scores between $F_0$ and its most relevant memory feature (first frame in $\mathbf{M}[F_{1:t}]$) have worse concentration on the correct object and become scattered in a longer memory bank. Therefore, we conclude that restricting the memory banks with a concise set of relevant features potentially benefits the decoding of VOS modules via more precise attention. 

\vspace{\secspace}
\section{Method of \ourmodel}
\label{sec:method}
\vspace{\secspace}

Motivated by our insight from the pilot study, we propose a straightforward approach highlighting a concise memory bank: restricting the memory with a constant frame number (Sec.~\ref{sec:restrict_memory}). We then explore the strategies to update the memory bank to constantly digest incoming features and remove obsolete frames (Sec.~\ref{sec:mem_update}). Finally, the restricted memory bank decreases the gap between the memory lengths across the training and inference stages. This enables previously overlooked techniques, and we propose a compelling example of temporal positional embedding (Sec.~\ref{sec:mem_temporal}). The overview of our method ``\ourmodel'' (``R'' for ``Restricted'') is in Fig.~\ref{fig:method}.

\vspace{\secspace}
\subsection{Restricting Memory Banks for VOS}
\label{sec:restrict_memory}
\vspace{\secspace}

\mypar{Design.} As indicated in our pilot study (Sec.~\ref{sec:pilot}), VOS modules have limited capability to process large quantities of features and thus benefit from a concise memory bank with less redundant information. To verify this in actual VOS systems, we develop the simple approach of \emph{restricting the memory bank to a fixed frame number}. In practice, a pre-defined small constant number $K$ is the maximum number of frames a memory bank can store, as shown in Fig.~\ref{fig:method}. The simplicity of our approach makes it a \emph{plug-and-play} enhancement for the existing VOS framework.

At an arbitrary frame $t$, we simplify the notation of the memory bank by denoting $\mathbf{M}[F_{0:t-1}]$ as $\mathbf{M}^t$, containing $K_t\leq K$ frames. A natural issue of bounded memory $\mathbf{M}^t$ is that $K_t$ can reach the limit $K$ at sufficiently large $t$, making the digestion of newly arriving features non-trivial, especially when the quality of information is vital for VOS, according to how we address degradation in the pilot study (Sec.~\ref{sec:pilot}). Our baseline adopts an intuitively simple yet effective approach (we explore better strategies in Sec.~\ref{sec:mem_update}): selecting the most reliable frame (frame 0) and temporally most relevant frames (closest frames). Formally, updating the memory bank is as below when $K_t = K$:
\begin{equation}
    \label{eq:mem_update_baseline}
    \mathbf{M}^{t+1} = \text{Concat}(\mathbf{M}_{0}^{t},\ \mathbf{M}_{2:K_t-1}^{t},\ F_t),
\end{equation}
where $\mathbf{M}_{2:K_t-1}^{t}$ and $F_t$ are the closest frames, and $\mathbf{M}_{1}^{t}$ is removed to create an available slot, as shown in Fig.~\ref{fig:method} (b).

\mypar{Discussion.} Our restricted memory is a revisit to previous methods~\cite{liang2020video, li2020fast}. However, we are distinct in emphasizing \emph{accuracy} instead of \emph{efficiency}. In addition, our \ourmodel also simplifies them~\cite{liang2020video, li2020fast, cheng2022xmem} by treating each frame as a constituent feature map instead of breaking it into smaller regions or pixels~\cite{cheng2022xmem}; thus, our strategy can directly apply to a wider range of models. Although more sophisticated strategies might further improve our accuracy, a simple approach is already effective (Sec.~\ref{sec:sota}).

\vspace{\secspace}
\subsection{Memory Update}
\label{sec:mem_update}
\vspace{\secspace}

Updating the incoming frames to the memory bank provides informative cues for VOS modules to decode. Although our baseline (Eqn.~\ref{eq:mem_update_baseline}) has already cooperated with the bounded memory bank, we investigate better methods for updating.

\mypar{Challenges of Memory Update.} As shown in our pilot study (Sec.~\ref{sec:pilot}), improving the conciseness of information heavily influences the decoding efficacy of VOS modules. Therefore, naive heuristics of random selection or keeping the latest frames are unreliable (as in Sec.~\ref{sec:ablation}, memory update analysis), since they fail to consider the relevance of frames (random) or suffer from drifting of knowledge (latest). To this end, we propose the principles that consider both \emph{relevant} prototypical features and \emph{fresh} incoming information from the latest frames.

\mypar{Memory Update Inspired by Multi-arm Bandits.} Our memory update problem can be stated as \emph{how to select and delete the most obsolete frame $k_d$ from $K$ candidates} to create slots for incoming features. Although not exactly identical, this problem analogizes \emph{multi-arm bandit}~\cite{katehakis1987multi}, which also concerns optimizing the reward by selecting from a fixed number of candidates. Its most inspiring insight for us is balancing the exploitation and exploration with the upper confidence bound (UCB) algorithm~\cite{auer2002using}, whose maximization objective $O_k$ for an option $k$ is as below,
\begin{equation}
    \label{eq:ucb}
    O_k = R_k + \sqrt{(2\log T)/t_k},
\end{equation}
where $R_k$ is option $k$'s average reward, $T$ is the total timestamps, and $t_k$ is the number of timestamps selecting $k$. When applying to our VOS, we re-define $R_k$ as the relevance of a frame for reliable VOS and consider $\sqrt{(2\log T)/t_k}$ as the freshness of memory, intuitively. Then, the deleted frame $k_d$ is chosen according to the smallest $O_{1:K}$. 
In practice, we define the relevance term $R_k$ using the attention scores between frame $\mathbf{M}_k^t$ and current VOS target $F_t$, to quantify the contribution of features from the memory. Under the context of transformers, we assume decoding the memory bank is as
\begin{equation}
\label{eq:memory_read}
\begin{split}
    F^D_t = \attn(& \attnQ = F_t, \attnK = \mem^t, \attnV = \mem^t),
\end{split}
\end{equation}
and assume that $\mathbf{S}^t$ is the scores (after softmax) between $F_t$ and $\mem^t$, computed inside the attention. Then, we treat the sum of scores as the relevance of a frame in the memory: $R_k=\text{sum}(S^t_{k})$, where $S^t_{k}$ is the slice of attention scores corresponding to $\mem^t_k$. Compared to XMem~\cite{cheng2022xmem}, which also uses attention scores for selection, our design differs in selecting at the frame level instead of the pixel level, which is simpler and already effective (as in Sec.~\ref{sec:ablation}).

As for the second term in UCB, $\sqrt{(2\log T)/t_j}$, we modify it by defining $t_j$ as the times a frame has stayed in the memory bank and $T$ as the sum of all the frames' staying time. This freshness term penalizes long-staying frames and allows refreshing from the latest information. Finally, $O_k$ combines it with the relevance term $R_k$ via a weight $\alpha$ balancing their numerical scales.

\vspace{-2mm}
\subsection{Memory with Temporal Awareness}
\label{sec:mem_temporal}
\vspace{\secspace}

\mypar{Motivation.} In addition to accommodating the decoding capability of VOS modules, restricting the memory bank systematically decreases the training-inference discrepancies in memory lengths. Specifically, the VOS algorithms are generally trained on short video clips with a few frames in the memory, while the videos are much longer during inference time. Therefore, the number of frames in the memory bank diverges more significantly without our restriction. 

Such temporal alignment between training and inference opens new opportunities for VOS. As a compelling example,  we introduce temporal positional embedding (PE) to enhance spatio-temporal reasoning. Specifically, we notice that previous approaches~\cite{yang2021aot, cheng2022xmem, cheng2021rethinking} overlook the order of frames in the memory, \emph{i.e.}, the temporal relationship among the frames are not explicitly considered, while spatial PE is widely adopted. Considering the vital role of orders in temporal modeling, which is commonly addressed with temporal PE in video-based tasks, we conjecture that the distinction of memory sizes between training and inference hinders previous methods from employing temporal PE.

\mypar{Design.} The objective of temporal PE is to embed explicit temporal awareness into memory and guide the attention in Eqn.~\ref{eq:memory_read}. Although restriction on the memory bank alleviates the training-inference shift, the challenges of temporal PE still exist: the optimal memory size $K$, though much smaller than expanding, can still be larger than the training-time memory size $K_{\text{train}}$; (2) the frames in the memory are varying from 1 to $K$. To address them, our solution is inspired by how ViT~\cite{dosovitskiy2020image} uses learnable PE and interpolation to address different image resolutions. Similarly, we initialize the PE according to $K_{\text{train}}$, denoted as $\widetilde{P}_{0:K_{\text{train}}-1}$, and the query $F_t$ having a dedicated PE $P_q$. Then, the temporal PE for the memory bank $\mathbf{M}^{t}_{0:K_t-1}$ is $P^t_{0:K_t-1}$.
\begin{equation}
    \label{eq:temporal_pe}
    P^t_{0:K_t-1} = \begin{cases} \widetilde{P}_{0:K_{t}-1},& K_t \leq K_{\text{train}}  \\
                \text{Interp}(\widetilde{P}_{0:K_{\text{train}}-1}, K_t), & K_t >  K_{\text{train}} \end{cases} 
\end{equation}
where ``$\text{Interp}(\cdot)$'' interpolates $\widetilde{P}_{0:K_{\text{train}}-1}$ to $K_t$ via nearest neighbor.
Finally, temporal PE enhances the original attention in Eqn.~\ref{eq:memory_read} by augmenting the key and values, identical to our conceptual illustration in Fig.~\ref{fig:method} (c):
\begin{equation}
\label{eq:memory_read_with_pe}
\begin{split}
    F^D_t = \text{Attn}(& \text{Q} = F_t + P_q,\\
    & \text{K} = \mathbf{M}^{t}_{0:K_t-1} + P^t_{0:K_t-1}, \\
    & \text{V} = \mathbf{M}^t_{0:K_t-1}).
\end{split}    
\end{equation}

The above design contains two critical choices. (1) We use the relative index $\{k=0,..., K_t-2\}$ inside the memory instead of the frame index $t$ to avoid the shift between training and inference. (2) Using learnable PE instead of Fourier features fits better to a limited training length, $K_{\text{train}}$.

\vspace{\secspace}
\section{Experiments}
\vspace{\secspace}
\subsection{Datasets and Evaluation Metrics}
\vspace{\secspace}
\mypar{VOST.} We primarily utilize the recent VOST~\cite{tokmakov2023breaking} dataset that concentrates on challenging object state changes. It curates over 700 videos covering diverse object state changes, \emph{e.g.}, changing appearance, occlusions, crowded objects, and fast motion. In VOST, the evaluation metrics are $\mathcal{J}$ and $\mathcal{J}_{\mathrm{tr}}$, resembling the average Jaccard over \emph{all the frames} and the harder \emph{last 25\% frames} corresponding to state changes.

\mypar{Long Videos Dataset.} We use the Long Videos dataset~\cite{liang2020video} to evaluate long-term understanding, similar to XMem~\cite{cheng2022xmem}. It contains 3 validation videos with more than 1k frames. $\jaccd$, $\contour$ (boundary F measure), and $\jaccd\& \contour$ (average of $\jaccd$, $\contour$) are considered for evaluation.

\mypar{LVOS.} We also experiment with the recent LVOS~\cite{hong2023lvos} dataset and include the results in \refsec{sec:result_lvos}. 

\mypar{Regular and Short Video Datasets.} YoutubeVOS~\cite{xu2018youtube} and DAVIS~\cite{pont20172017, perazzi2016benchmark} are two earlier datasets with \emph{shorter} duration and \emph{easier} scenarios compared with VOST. In this paper, we use them as the pretraining datasets for VOST and the Long Videos dataset following standard practice~\cite{tokmakov2023breaking, cheng2022xmem}, and conduct 
 analysis in addition to the challenging datasets.

\vspace{\secspace}
\subsection{Baselines and Implementation Details}
\label{sec:baseline_impl_detail}
\vspace{\secspace}

Our proposed \ourmodel is a simple and plug-and-play enhancement for the VOS framework. Without loss of generality, we select AOT~\cite{yang2021aot} and DeAOT~\cite{yang2022decoupling} as the main baseline because of its top performance on VOST (as in \reftab{tab:vost_sota}) and simplicity. It adopts ResNet-50~\cite{he2016deep} as its encoder and a specially designed ``long short term-transformer'' (LSTT) as its decoder. For the memory bank, the original AOT digests the latest frame and expands the memory continuously, while \ourmodel restricts its size to 8 frames. We also employ \ourmodel on other VOS methods in addition to AOT. More details on models and implementation in \refsec{sec:supp_implementation_details}.

\begin{table}[t]
\centering
\resizebox{0.61\linewidth}{!}{
\begin{tabular}{l@{\hspace{8mm}}c@{\hspace{6mm}}c@{\hspace{6mm}}}
\toprule
 & $\mathcal{J}_{\mathrm{tr}}$ & $\mathcal{J}$  \\ \midrule
OSMN Match~\cite{yang2018efficient} & 7.0 & 8.7 \\
OSMN Tune~\cite{yang2018efficient} & 17.6 & 23.0 \\
CRW~\cite{jabri2020space} & 13.9 & 23.7 \\
CFBI~\cite{yang2020collaborative} & 32.0 & 45.0 \\
CFBI+~\cite{yang2021collaborative} & 32.6 & 46.0 \\
XMem~\cite{cheng2022xmem} & 33.8 & 44.1 \\
HODOR Img~\cite{athar2022hodor} & 13.9 & 24.2 \\
HODOR Vid~\cite{athar2022hodor} & 25.4 & 37.1 \\
AOT~\cite{yang2021aot} & 36.4 & 48.7 \\
\midrule
AOT$^\Psi$                   & 37.0 & 49.2 \\
AOT$^\Psi$ + \ourmodel (Ours) & 39.8 & 50.5 \\
\midrule
DeAOT$^\Psi$                  &  37.6 & 50.1 \\
DeAOT$^\Psi$ + \ourmodel (Ours) & \textbf{40.4} & \textbf{51.8} \\
\bottomrule
\end{tabular}
}
\vspace{-3mm}
\caption{Comparisons with previous methods on VOST~\cite{tokmakov2023breaking}. Our \ourmodel shows advantages on both overall quality ($\mathcal{J}$) and addressing object state changes ($\mathcal{J}_{\mathrm{tr}}$). (If not specified, the results are from VOST's implementation, $\Psi$ denotes our implementation.)}
\vspace{-4mm}
\label{tab:vost_sota}
\end{table}

\vspace{\secspace}
\subsection{State-of-the-art Comparisons}
\label{sec:sota}
\vspace{\secspace}

\mypar{VOST.} In \reftab{tab:vost_sota}, we compare \ourmodel with previous methods on VOST. Our approach establishes new state of the art on this challenging benchmark with a significant improvement. Notably, our simple strategy increases the VOS quality for the whole video ($\jaccd$) and maintains better robustness for the state-changing frames ($\jaccd_{\mathrm{tr}}$). This is especially clear when compared to AOT~\cite{yang2021aot}: the improvement is over $\sim$3\% with our plug-and-play modifications.

\begin{table}[t]
\centering
\resizebox{0.77\linewidth}{!}{
\begin{tabular}{l@{\hspace{8mm}}c@{\hspace{6mm}}c@{\hspace{6mm}}c@{\hspace{6mm}}}
\toprule
                    & $\jaccd \& \contour$  & $\jaccd$ & $\contour$ \\ \midrule
CFBI~\cite{yang2020collaborative} & 53.5    & 50.9     & 56.1  \\
CFBI+~\cite{yang2021collaborative}& 50.9    & 47.9     & 53.8  \\
STM~\cite{oh2019video} & 80.6    & 79.9     & 81.3  \\
MiVOS~\cite{cheng2021modular} & 81.1    & 80.2     & 82.0  \\
AFB-URR~\cite{liang2020video} & 83.7    & 82.9     & 84.5  \\
STCN~\cite{cheng2021rethinking} & 87.3    & 85.4     & 89.2  \\
XMem~\cite{cheng2022xmem}         & 89.8    & 88.0     & 91.6  \\
AOT~\cite{yang2021aot}            & 84.3    & 83.2     & 85.4  \\
\midrule
AOT$^\Psi$                        & 86.7    & 85.5     & 87.9  \\
AOT$^\Psi$  + \ourmodel (Ours)   & 90.3    & 88.5 & 92.1  \\
\midrule
DeAOT$^\Psi$                     & 89.4   & 87.4      & 91.4  \\
DeAOT$^\Psi$ + \ourmodel (Ours)  & \textbf{91.5}   & \textbf{89.8}      & \textbf{93.3}   \\
\bottomrule
\end{tabular}
}
\vspace{-3mm}
\caption{Comparison with previous methods on Long Videos dataset~\cite{liang2020video}. For both baselines of AOT and DeAOT, our \ourmodel shows significant improvement. (Without mention, the results are from XMem~\cite{cheng2022xmem}, $\Psi$ denotes our implementation.)}
\vspace{-5mm}
\label{tab:long_video_sota}
\end{table}

\mypar{Long Videos Dataset.} As our \ourmodel limits memory capacity, a natural suspicion is that our memory bank performs worse in storing information and struggles with long-term modeling. However, our comparison in \reftab{tab:long_video_sota} shows the opposite. On the Long Videos dataset, our \ourmodel not only improves upon the baseline AOT and DeAOT models but also outperforms the state of the art XMem~\cite{cheng2022xmem} model, which utilizes specially designed hierarchical memory banks and memory manipulation operators. Therefore, this further supports our insight on keeping a concise memory bank to accommodate the limited capability of VOS modules to address expanding memory banks. 

\vspace{\secspace}
\subsection{Ablation Studies}
\label{sec:ablation}
\vspace{\secspace}

\mypar{Effect of \ourmodel Components.} We analyze each \ourmodel component respect to AOT and DeAOT baselines, as in Table~\ref{tab:ablation}. \textbf{(1)} \textit{Restricting memory banks.} The most important insight from our pilot study (Sec.~\ref{sec:pilot}) is to maintain a concise memory bank with relevant information, which motivates our revisit of restricting memory banks (Sec.~\ref{sec:restrict_memory}). According to Table~\ref{tab:ablation} (row 1 and 2), a bounded memory bank leads to significant enhancement in the long and complex VOST videos. \textbf{(2)} \textit{Temporal positional embedding.} In Table~\ref{tab:ablation}, we illustrate that adding positional embedding (Sec.~\ref{sec:mem_temporal}) greatly benefits the spatio-temporal modeling, especially the harder $\mathcal{J}_{\mathrm{tr}}$ for state changes. \textbf{(3)} \textit{Memory update.} We refresh the memory banks by balancing the relevance and freshness of frames (Sec.~\ref{sec:mem_update}), inspired by the UCB algorithm~\cite{auer2002using}. In rows 4 and rows 5 of Table~\ref{tab:ablation}, such a strategy effectively boosts the overall performance.

\begin{table}[t]
\centering
\resizebox{0.67\linewidth}{!}{
\begin{tabular}
{c@{\hspace{2mm}}|c@{\hspace{2mm}}c@{\hspace{2mm}}c@{\hspace{2mm}}|c@{\hspace{3mm}}c@{\hspace{3mm}}|c@{\hspace{3mm}}c@{\hspace{3mm}}}
\toprule
\multirow{2}{*}{ID} & \multirow{2}{*}{RM} & \multirow{2}{*}{TPE} & \multirow{2}{*}{MU} & \multicolumn{2}{c|}{AOT} & \multicolumn{2}{c}{DeAOT} \\
\cline{5-6} \cline{7-8}
& & &  & $\jaccd_{\mathrm{tr}}$ & $\jaccd$ & $\jaccd_{\mathrm{tr}}$ & $\jaccd$ \\ 
\midrule
   1  &  \multicolumn{3}{c|}{Baseline} & 37.0 & 49.2 & 37.6 & 50.9 \\ \midrule
 2 & \cmark &        &        & 38.7 & 50.3 & 38.8 & 51.0 \\
 3 & \cmark & \cmark &        & 39.7 & 50.3 & 40.0 & 51.7 \\
 4 & \cmark &        & \cmark & 39.4 & 50.3 & 39.0 &  51.4  \\
 5 & \cmark & \cmark & \cmark & \textbf{39.8} & \textbf{50.5} & \textbf{40.4} & \textbf{51.8} \\
\bottomrule
\end{tabular}
}
\vspace{-3mm}
\caption{Ablation studies of \ourmodel on VOST. Starting from the AOT and DeAOT baselines, all of the components improve the performance, especially the harder object state-changing frames ($\mathcal{J}_{tr}$). \textbf{RM}: restricting memory banks. \textbf{TPE}: temporal positional embedding. \textbf{MU}: memory update with the UCB algorithm. }
\vspace{-4mm}
\label{tab:ablation}
\end{table}
\begin{table}[t]
\centering
\resizebox{0.55\linewidth}{!}{
\begin{tabular}
{
l@{\hspace{3mm}}|
l@{\hspace{3mm}}|
c@{\hspace{3mm}}c@{\hspace{3mm}}
}
\toprule
     Method & Variant  & $\jaccd_{tr}$ & $\jaccd$ 
     \\
\midrule
\multirow{5}{*}{Remove} & 0$^{th}$  & 35.9  & 48.9 
\\
& 1$^{st}$   & 38.7 & 50.3 
\\
& Middle  & 38.3 & 50.2 
\\
& Latest  & 35.7 & 48.5 
\\
& Random & 38.0 & 50.0 
\\
\midrule
\multirow{2}{*}{UCB}
& Relev & 39.1 & 50.1 
\\
& Relev + Fresh  & \textbf{39.4} & \textbf{50.3}  
\\
\bottomrule
\end{tabular}
}
\vspace{-3mm}
\caption{Ablation study of different memory updating strategies on VOST. We analyze deleting a frame in the memory based on heuristics (``Remove'') or guided by the relevance and freshness of the UCB algorithm (``UCB''). Our final memory updating strategy using both relevance and freshness achieves the best performance.}
\vspace{-4mm}
\label{tab:ablation_mem_update}
\end{table}

\begin{figure}
    \centering
    \includegraphics[width=0.83\linewidth]{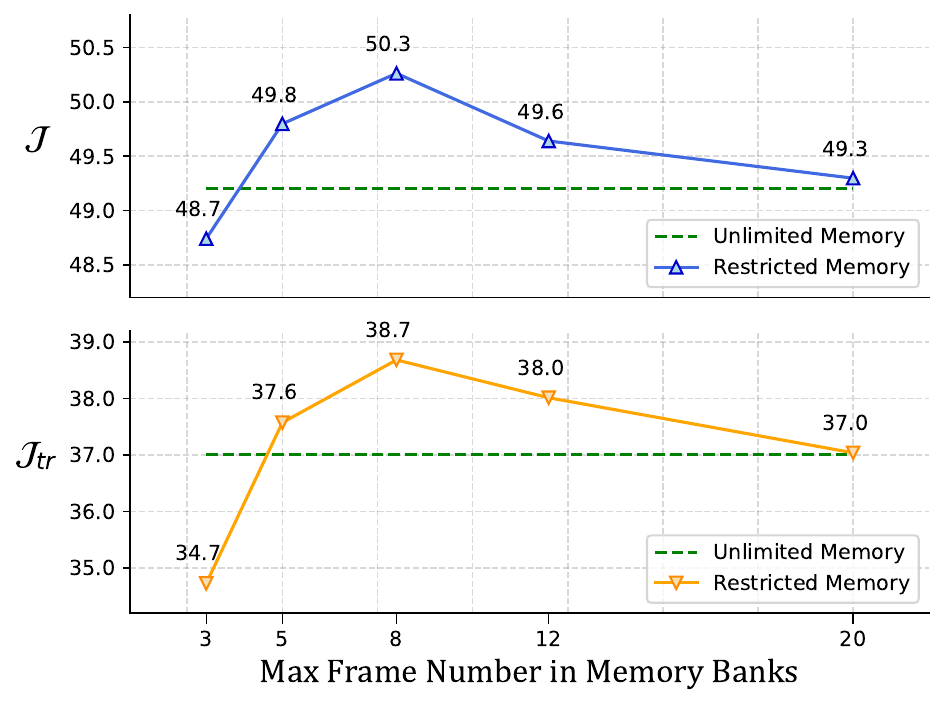}
    \vspace{-3mm}
    \caption{Impact of memory bank size on VOS, tested on VOST. With more frames in the restricted memory, the accuracy first increases and then decreases until it approximates unrestricted memory. This supports the limited deciphering capability of VOS modules and our insight into restricting memory banks.}
    \vspace{-4mm}
    \label{fig:ablation_mem_len}
\end{figure}

\mypar{Analysis on Frame Numbers of Memory Banks.} We verify a direct implication of our insight: an expanding memory bank elevates the difficulty of VOS modules to decode information. Specifically, we observe the VOS accuracy under various sizes of memory banks. To avoid the influence of hyper-parameter tuning, we utilize the baseline memory update strategy in Sec.~\ref{sec:restrict_memory}. As in Fig.~\ref{fig:ablation_mem_len}, the performance first improves from richer information. Then both $\mathcal{J}$ and $\mathcal{J}_{\mathrm{tr}}$ decrease when the length of memory exceeds the capability of learned AOT modules, until they become similar to unrestricted memory. Consequently, these results directly support our insight of restricting memory banks. 

\begin{figure*}
    \centering
    \includegraphics[width=0.95\textwidth]{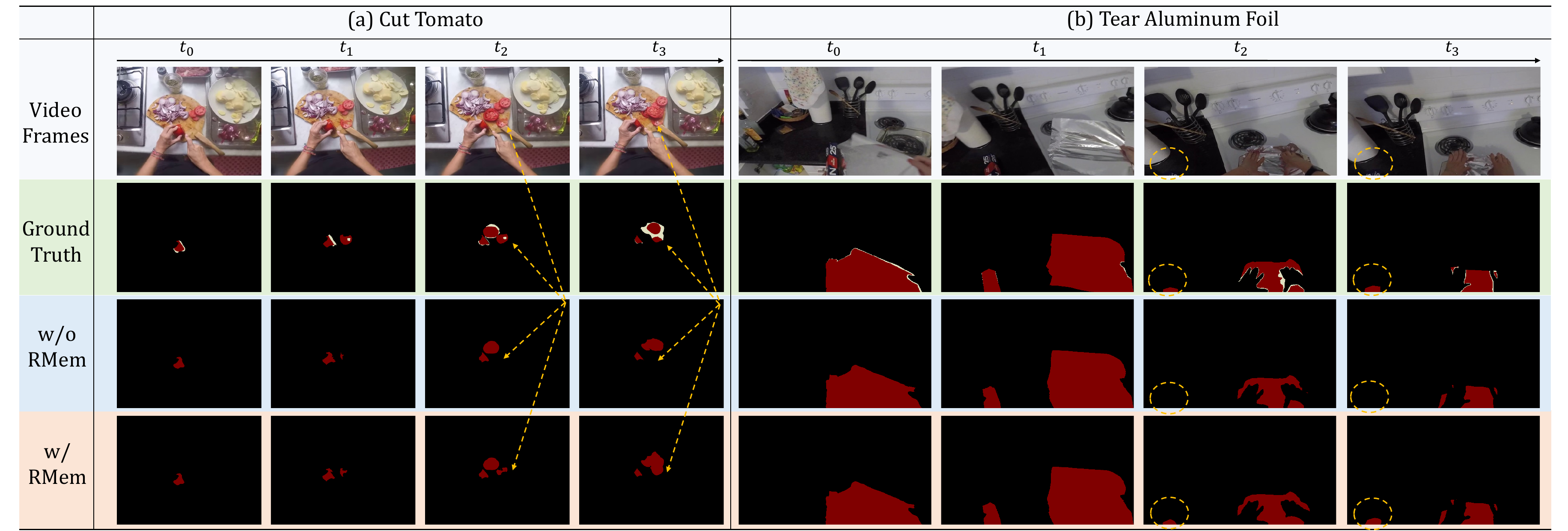}
    \vspace{-3mm}
    \caption{(Best viewed zoom-in with color.) Qualitative VOS results for object state changes on VOST~\cite{tokmakov2023breaking}. We provide two examples showing the challenges of object state changes, including slicing, occlusions, distraction from similar objects (other tomatoes), and shape changes. For both scenarios, using \ourmodel shows advantages in robustly maintaining the masks of the target objects, as highlighted. (White pixels are annotated by VOST denoting ``ignored'' regions for evaluation, which are hard and ambiguous even for human annotators.)}
    \vspace{-5mm}
    \label{fig:visualization}
\end{figure*}

\mypar{Memory Update Analysis.}  Maintaining an informative memory bank is critical for the VOS accuracy, and we propose a UCB-inspired algorithm in Sec.~\ref{sec:mem_update}. Table~\ref{tab:ablation_mem_update} analyzes the key intuition and design choices with AOT. \textbf{(1)} The initial frame is critical in keeping the provided ground-truth information: removing the 0-th frame leads to an accuracy drop, and is more profound when scenarios are complex (VOST). \textbf{(2)} Guaranteeing the freshness of information is critical, where removing the latest frame leads to the worst accuracy. \textbf{(3)} Randomly removing frames performs surprisingly well but is still worse than our baseline (removing the 1$^{st}$ frame, in Sec.~\ref{sec:restrict_memory}). \textbf{(4)} Using attention scores to reflect the relevance better removes redundant features (``Relev''), and it is further enhanced with the freshness term, where freshness is especially effective to avoid frames from staying long time in the memory bank, supported by the Long Video dataset. Finally, the best strategy is our UCB-inspired algorithm combining relevance and freshness.

\begin{table}[t]
\centering
\resizebox{0.7\linewidth}{!}{
\begin{tabular}{l@{\hspace{8mm}}|c@{\hspace{6mm}}c@{\hspace{6mm}}}
\toprule
Method & $\mathcal{J}_{tr}$ & $\mathcal{J}$  \\ \midrule
AOT & 37.0 & 49.2 \\
AOT + RM & 38.7 & 50.3 \\
\midrule
AOT + SinCos PE & 37.2 & 48.3 \\
AOT + Learnable PE & 36.7 & 49.4 \\
\midrule
AOT + RM + SinCos PE & 37.9 & 48.9 \\
AOT + RM + Learnable PE & \textbf{39.7} & \textbf{50.3} \\ \bottomrule
\end{tabular}
}
\vspace{-3mm}
\caption{Comparison of temporal PE strategies on VOST. Based on restricted memory (``RM''), our learnable temporal PE (``Learnable'') is better than using high-frequency Fourier features (``SinCos''). Notably, restricting memory is essential for PE.}
\vspace{-5mm}
\label{tab:ablation_pe}
\end{table}

\mypar{Temporal Positional Embedding Strategies.} We introduce using learnable temporal PE to address the varied frames in the memory banks of VOS in Sec.~\ref{sec:mem_temporal}. In \reftab{tab:ablation_pe}, we analyze another PE strategy of encoding the index into high-frequency features with SinCos functions and find it performs worse. This is because SinCos is commonly used in scenarios of a large number or continuous space of coordinates (\emph{e.g.}, NeRF~\cite{mildenhall2020nerf}), while learnable embeddings can better handle a small number of slots (\emph{e.g.}, ViT~\cite{dosovitskiy2020image}), as in the limited memory length during the VOS training. Furthermore, we highlight that temporal PE requires restricted memory to function well because of better training-inference temporal alignment in memory lengths. This supports our intuition in Sec.~\ref{sec:mem_temporal} and suggests the emerging opportunities from restricting memory banks.

\mypar{Analysis on Regular and Short Video Benchmarks.} We highlight the improvement on long and complex VOS datasets, but we also supplement our analysis on the regular and short video dataset DAVIS2017. As in Table~\ref{tab:ablation_short_video}, our \ourmodel has relatively the same performance but effectively improves the efficiency. Compared with our improvement on VOST and the Long Video dataset, we conjecture that the learned VOS modules (AOT and DeAOT) are already capable of handling shorter video duration and less complicated scenarios, even without our concise memory banks. Additionally, this potentially explains that previous studies exploring restricting memory banks~\cite{li2013video, liang2020video} have not explicitly discovered its benefits, \emph{probably due to not considering longer and more challenging datasets like VOST}.

\begin{table}[t]
\centering
\resizebox{0.9\linewidth}{!}{
\begin{tabular}
{
l@{\hspace{3mm}}|
c@{\hspace{3mm}}c@{\hspace{3mm}}c@{\hspace{3mm}}|
c@{\hspace{3mm}}c@{\hspace{3mm}}}
\toprule
Method & $\jaccd \& \contour$ & $\jaccd$ & $\contour$  & Max Mem $\downarrow$ & FPS \\
\midrule
AOT                    & 85.2 & 82.5 & 87.9 & 4.46G & 13.67 \\
AOT + \ourmodel (Ours) & 85.2 & 82.4 & 88.0 & 2.34G & 15.57 \\
\midrule

DeAOT                  & 85.2 & 82.3 & 88.1 & 2.24G & 25.11 \\
DeAOT + \ourmodel (Ours)& \textbf{85.3} & \textbf{82.3} & \textbf{88.2} & \textbf{1.53G} & \textbf{27.42} \\
\bottomrule
\end{tabular}
}
\vspace{-3mm}
\caption{\ourmodel maintains the accuracy on DAVIS2017 while being more efficient, indicating that \ourmodel can be generally applied, not limited to challenging scenarios. This also aligns with the prior works and suggests that not having demanding datasets was potentially why the \emph{accuracy} benefits of memory restriction were not clearly revealed previously.}
\vspace{-6mm}
\label{tab:ablation_short_video}
\end{table}

\vspace{\secspace}
\subsection{Qualitative Results}
\vspace{\secspace}

We visualize on two representative videos from VOST~\cite{tokmakov2023breaking} that require robust spatio-temporal reasoning in Fig.~\ref{fig:visualization}. Video (a) is the kitchen behavior of cutting a tomato into slices, and it illustrates the challenges of splitting objects, occlusions from hands, and visual distraction from other tomatoes. Without our \ourmodel, the baseline AOT model fails to maintain the masks for the separated tomato slice, while using \ourmodel correctly remembers this slice at the later stage of the video (columns 3 and 4). Such regions are highlighted with the yellow arrows. The other video (b) illustrates another difficulty of object shape transformation and splitting between the box and the aluminum. Although the baseline model without \ourmodel can correctly segment the box at the beginning of splitting (column 2), it gradually loses track of the box and can only concentrate on the dominant object. However, our model enhanced with \ourmodel robustly segments the small regions of the box, indicating that its attention association with relevant historical frames is still stable because of our restricted memory. Therefore, we conclude that the quantitative results reveal the difficulties of object state changes and support the effectiveness of our approach.  

\vspace{-3mm}
\section{Conclusion}
\label{sec:conclusion}
\vspace{-3mm}

This paper reveals the drawbacks of expanding memory banks, a conventional design in VOS. Our insight stems from a novel ``memory deciphering'' analysis, which suggests that the redundant information in growing memory banks confuses the attention of VOS modules and elevates the difficulty of feature decoding. Then, we propose the simple enhancement for VOS named \ourmodel. At its core is restricting the size of memory banks, accompanied by UCB-inspired memory update strategies and temporal positional embedding to enhance spatio-temporal reasoning. Extensive evaluation on the recent challenging datasets, including VOST and the Long Videos dataset, supports our insight and effectiveness of \ourmodel.

\mypar{Limitations and Future Work.} Our paper prioritizes the analysis of memory banks and illustrates our insight with a straightforward approach. Therefore, interesting future work is to combine the intuition from more sophisticated methods, such as XMem~\cite{cheng2022xmem}. Furthermore, our exploration mainly adapts memory banks to cooperate with the capability of VOS modules, while how to improve the decoding ability of VOS modules for a huge memory bank is the alternative direction and interesting future work.

{
    \small
    \bibliographystyle{ieeenat_fullname}
    \bibliography{main}
}
\newpage
\section*{Appendix}

\renewcommand\thesection{\Alph{section}}
\renewcommand\thetable{\Alph{table}}
\renewcommand\thefigure{\Alph{figure}}
\renewcommand\theequation{\Alph{equation}}
\setcounter{section}{0}
\setcounter{table}{0}
\setcounter{figure}{0}

Our appendix cover additional analysis, implementation details, and discussion as below:
\begin{enumerate}[leftmargin=*, noitemsep, nolistsep, label=(\Alph*)]
    \item \textbf{Demo Video.} We provide a demo video at \href{https://youtu.be/mFjGSPXmXdA}{https://youtu.be/mFjGSPXmXdA} showing multiple challenging VOS examples (Sec.~\ref{sec:supp_video}).
    \item \textbf{Implementation details.} We explain the detailed model architectures and the procedures for training and inference (Sec.~\ref{sec:supp_implementation_details}).
    \item \textbf{Additional ablation studies.} This section provides more analysis and experimental results (Sec.~\ref{sec:supp_ablation}).
    \item \textbf{Addition discussion on limitations and future work.} We offer a more detailed discussion of the limitations and potential future directions (Sec.~\ref{sec:supp_limitations}).
\end{enumerate}

\section{Demo Video}
\label{sec:supp_video}

In \href{https://youtu.be/mFjGSPXmXdA}{https://youtu.be/mFjGSPXmXdA}, we provide four qualitative comparison examples between the baseline models (AOT~\cite{yang2021aot} and DeAOT~\cite{yang2022decoupling}) and our \ourmodel, with the object state changes from both VOST~\cite{tokmakov2023breaking} and the Long Videos dataset~\cite{liang2020video}. Notably, these examples illustrate four challenging scenarios: (1) \textbf{Object ambiguity}: objects have similar appearances; (2) \textbf{Slicing}: an object is cut into multiple slices; (3) \textbf{Appearance changes}: an object has changed its shape and appearances, leading to incorrect VOS masks. (4) \textbf{Sudden shape changes}: the viewpoint changes quickly and causes variation in shapes of the target object. The four examples demonstrate that \ourmodel effectively improves the spatio-temporal reasoning of VOS. 

\section{Implementation Details}
\label{sec:supp_implementation_details}

We describe the outline of implementation of AOT~\cite{yang2021aot} and DeAOT~\cite{yang2022decoupling} baselines in \refsec{sec:baseline_impl_detail}. This section provides in-depth details of the implementation and the configuration of \ourmodel.

\subsection{Model Architecture}

AOT and DeAOT share the common architecture of the memory-based VOS framework. As conceptualized in \refeq{eq:encoder_decoder}, we disassemble the VOS framework into the modules of an encoder $\mathbf{E}(\cdot)$ encoding images into feature maps, a decoder $\mathbf{D}(\cdot)$ extracting information from the memory bank, and a segmentation head translating the output from decoder into masks. Please note that we have additionally decoupled the segmentation head from the decoder for clarity, compared with \refeq{eq:encoder_decoder}.

\mypar{Encoder.} Identical to VOST~\cite{tokmakov2023breaking}, we adopt ResNet-50~\cite{he2016deep} as the encoder, which achieves competitive performance while efficient enough to operate on Long Videos. The multiple stages in the ResNet encoder produce 3 levels of feature maps $\{F^4, F^8, F^{16}\}$ with 1/4, 1/8, and 1/16 the resolution of the original input image, respectively. Following the practice of AOT and DeAOT, the deepest feature map $F^{16}$ is the input to the decoder for memory reading, and $\{F^4, F^8\}$ are provided to the segmentation head as input for predicting high-quality masks.

\mypar{Decoder.} AOT and DeAOT utilize a specially-designed transformer~\cite{vaswani2017attention} to conduct associative memory reading, named ``Long Short-term Transformer'' (LSTT). LSTT comprises three consecutive transformer layers to enhance features in the current frame with the memory bank. Adopting the same notations as \refeq{eq:encoder_decoder}, we conceptually illustrate this process as Eqn.~\ref{eq:lstt_layers}:

\begin{equation}
\label{eq:lstt_layers}
\begin{split}
    F^{(l+1)}_t = \attn(& \attnQ = F_t^{(l)}, \\ & \attnK = \mem^{(l)}[F_{0:t-1}], \\ & \attnV = \mem^{(l)}[F_{0:t-1}]),
\end{split}
\end{equation}
where the superscript $(l)$ denotes the layer index of LSTT, ranging from 0 to 2. After the above process, We keep the implementation details identical to the original AOT and DeAOT. Please refer to them for more detailed configuration. Finally, the output feature $F^{(3)}_t$ replaces the feature map $F^{16}$ from the encoder not enhanced with spatio-temporal information.

\mypar{Segmentation Head.} To maintain high-resolution segmentation masks, the segmentation process involves a feature pyramid network (FPN)~\cite{lin2017feature}. It accepts $F^{(4)}_t$ as the input feature, uses $\{F^8, F^{16}\}$ as shortcut inputs, and up-samples them via the combination of a convolutional layer and a bi-linear up-sampling layer.

\mypar{Temporal Positional Embedding.} We introduce temporal positional embedding (TPE) in \refsec{sec:mem_temporal} to enhance the spatio-temporal reasoning ability of models. In practice, we initialize end-to-end learnable embeddings with the same number to the memory length during the training time (\emph{e.g.}, 4 in VOST) and the same dimension to the feature $F_t$, marking the PE of each place in the memory bank. For simplicity, the three LSTT layers in Eqn.~\ref{eq:lstt_layers} share the same set of TPE.  

\subsection{Training}
\label{sec:supp_training}

\mypar{Loss Functions.} Our training procedure utilizes the same loss functions as AOT and DeAOT: the combination of bootstrapped cross-entropy loss and soft Jaccard loss~\cite{nowozin2014optimal}. Both loss terms are averaged 1:1 as the final loss value.

\mypar{VOST.} The training on VOST~\cite{tokmakov2023breaking} follows the original practice of VOST's authors, where the models are fine-tuned on VOST with pretrained weights from DAVIS2017~\cite{pont20172017} and Youtube2019~\cite{xu2018youtube}. As VOST highlights spatio-temporal modeling, we follow the authors' implementation of AOT by using a long sequence length of 15 frames during training and this accordingly enables 4 frames in the memory bank. It leverages exponential moving averages (EMA) for parameter updates to stabilize the training process. The whole training process uses AdamW~\cite{kingma2014adam, loshchilov2017decoupled} optimizer, and lasts 20,000 steps with a batch size of 8, on 4$\times$A40 GPUs. The initial learning rate is $2\times 10^{-4}$ and it gradually decays to $2\times 10^{-5}$ according to a polynomial pattern~\cite{yang2020collaborative}. To avoid overfitting, we set the learning rate of the encoder as 0.1 of the other components. The weight decay is 0.07, which is also identical to AOT and DeAOT.

\mypar{Long Videos Dataset.} Following the standard practice~\cite{cheng2022xmem, liang2020video}, we first train the AOT and DeAOT models on the DAVIS2017~\cite{pont20172017} and YoutubeVOS2019 dataset~\cite{xu2018youtube}, then conduct inference on the Long Videos dataset~\cite{liang2020video}. However, to support the training of positional embedding, we extend the length of training samples from the original 5 frames to 9 frames, to support 4 frames in the memory banks during the training time. Please note that we also re-train the baselines under the same setup to ensure a fair comparison. The training procedure leverages the similar optimization setting as described above for the VOST dataset, including the AdamW~\cite{kingma2014adam, loshchilov2017decoupled} optimizer, weight decay of 0.07, polynomial learning rate decay~\cite{yang2020collaborative} from $2\times 10^{-4}$ to $2\times 10^{-5}$, 0.1 scaling of the encoder learning rate, and EMA parameter updates. The only difference from VOST is training 100,000 steps with a batch size of 16, following the implementation of the original AOT and DeAOT on DAVIS2017 and YoutubeVOS2019 datasets.

\subsection{Inference.}

\mypar{VOST.} Instead of appending features into memory at \emph{a fixed frequency of 5 frames}, the authors of VOST developed a different strategy than on DAVIS2017 and YoutubeVOS2019 to address the CUDA memory issue caused by higher resolution and longer video duration: the memory bank is bounded by 30 frames and the frequency of updating memory banks is accordingly $L/30$, where $L$ is the length of the video. For our \ourmodel, we follow the frequency of memory updates set by VOST, but bounds the size of memory banks to 9 frames, which is significantly smaller than the original cap of 30 frames. Therefore, our \ourmodel needs to update the memory banks by removing the obsolete frames, and we describe the details of memory update in Sec.~\ref{sec:supp_detail_mem_update} below.

\mypar{Long Videos Dataset.} When comparing to the other approaches on the Long Videos dataset (\reftab{tab:long_video_sota}), we primarily rely on the VOS performance evaluated by XMem~\cite{cheng2022xmem}. However, we re-implement the baselines of AOT and DeAOT for a fair comparison with \ourmodel, since XMem has not released the code for evaluating both methods. Notably, \emph{our re-implementation achieves better performance} compared to XMem's reported numbers. In practice, we determine the frequency of updating memory banks by $L/30$ to avoid CUDA memory issues, which is similar to the inference procedure on VOST. Our \ourmodel shares the same inference setting as baseline, only restricting the memory bank size to 8 frames. Then, the memory update strategy is identical to VOST, as described in Sec.~\ref{sec:supp_detail_mem_update}.

\subsection{Memory Update}
\label{sec:supp_detail_mem_update}

As is described in \refsec{sec:mem_update}, our \ourmodel balances the relevance and freshness of frames in the memory bank using our algorithm inspired by UCB~\cite{auer2002using}.

\mypar{Relevance.} As mentioned in \refsec{sec:mem_update}, we use the attention scores from the transformers in the decoder \refeq{eq:memory_read} to reflect the relevance of a memory frame $R_k$. Since the LSTT decoder in AOT and DeAOT has three transformer layers, we intuitively select the attention scores from the 0-th transformer because it is closest to the original image embeddings $F_t$ and memory features $\mathbf{M}^t$ (ablation in Sec.~\ref{sec:supp_ablation_atten_score}). To stabilize the relevance term and avoid fluctuations, we further apply the moving average technique to the relevance term. Suppose $R_k^{'}$ denotes the relevance values of a memory frame $k$ derived from the latest timestamp, the consequent relevance term $R_k$ is updated via:
\begin{equation}
    R_k \xleftarrow{} (1 - \lambda) R_k^{'} + \lambda R_k,
\end{equation}
where we set $\lambda=0.8$ for both VOST and the Long Videos dataset. As we have noticed, using moving average for stabilization is a common technique for VOS on long videos, such as in AFB-URR~\cite{liang2020video}.

\mypar{Freshness.} To balance the numerical scales of the relevance and freshness terms, we slightly modify \refeq{eq:ucb} as below,
\begin{equation}
    \label{eq:balance_rel_fresh}
    O_j = R_j + \alpha \sqrt{\frac{\log T}{t_j + B}},
\end{equation}
where $B$ smooths the numerical ranges of the freshness term, and $\alpha$ controls the individual contribution of relevance and freshness. In practice, we set $B=8$ and $\alpha=1.5$ for both VOST and the Long Videos dataset. Detailed ablation studies on the values of $\alpha$ are illustrated in Sec.~\ref{sec:supp_balance_rel_fresh}.

\section{Supplemental Ablation Studies}
\label{sec:supp_ablation}

\subsection{Memory Update on the Long Videos Dataset}

We analyze the memory update strategies on the Long Videos dataset~\cite{liang2020video} using our AOT baseline in Table~\ref{tab:ablation_mem_update_long_videos}, in addition to the analysis on VOST~\cite{tokmakov2023breaking} (\reftab{tab:ablation_mem_update}). \textbf{(1)} Notably, we observe consistent improvement from our UCB-inspired memory update strategy combining both relevance and freshness of frames in the memory. \textbf{(2)} Similar to the results on VOST, our baseline of removing the 1-st frame in the memory has competitive performance but is inferior to our final UCB-inspired strategy. \textbf{(3)} The analysis in Table~\ref{tab:ablation_mem_update_long_videos} also reveals several intriguing differences between the Long Videos dataset and VOST. Specifically, VOST highly relies on the relevance of frames and the reliable information from the 0-th frames because of its complexity in scenarios, while the Long Videos dataset highlights the utility of freshness of frames as a consequence of extremely long video duration.

\begin{table}[h]
\centering
\resizebox{0.8\linewidth}{!}{
\begin{tabular}
{
l@{\hspace{3mm}}|
l@{\hspace{3mm}}|
c@{\hspace{3mm}}c@{\hspace{3mm}}c@{\hspace{3mm}}
}
\toprule
     Method & Variants & $\jaccd \& \contour$ & $\jaccd$ & $\contour$ 
     \\
\midrule
\multirow{5}{*}{Remove} & 0$^{th}$ & 88.1 & 86.3 & 89.9 
\\
& 1$^{st}$  & 88.3 & 86.6 & 90.1  
\\
& Middle & 86.6 & 85.5 & 87.9  
\\
& Latest & 85.4 & 84.1 & 86.7  
\\
& Random & 87.7 & 86.6 & 88.9 
\\
\midrule
\multirow{2}{*}{UCB}
& Relev & 86.9 & 85.4 & 88.3 
\\
& Relev + Fresh  & \textbf{89.5} & \textbf{87.8} & \textbf{91.2}  
\\
\bottomrule
\end{tabular}
}
\vspace{-3mm}
\caption{Ablation study of different memory updating strategies on the Long Videos dataset, in addition to VOST (\reftab{tab:ablation_mem_update}). We analyze deleting a frame in the memory based on heuristics (``Remove'') or guided by the relevance and freshness of the UCB algorithm (``UCB''). Our final memory updating strategy using both relevance and freshness achieves the best performance.}
\vspace{-4mm}
\label{tab:ablation_mem_update_long_videos}
\end{table}

\subsection{Balancing Relevance and Freshness}
\label{sec:supp_balance_rel_fresh}

\begin{figure}
    \centering
    \includegraphics[width=0.95\linewidth]{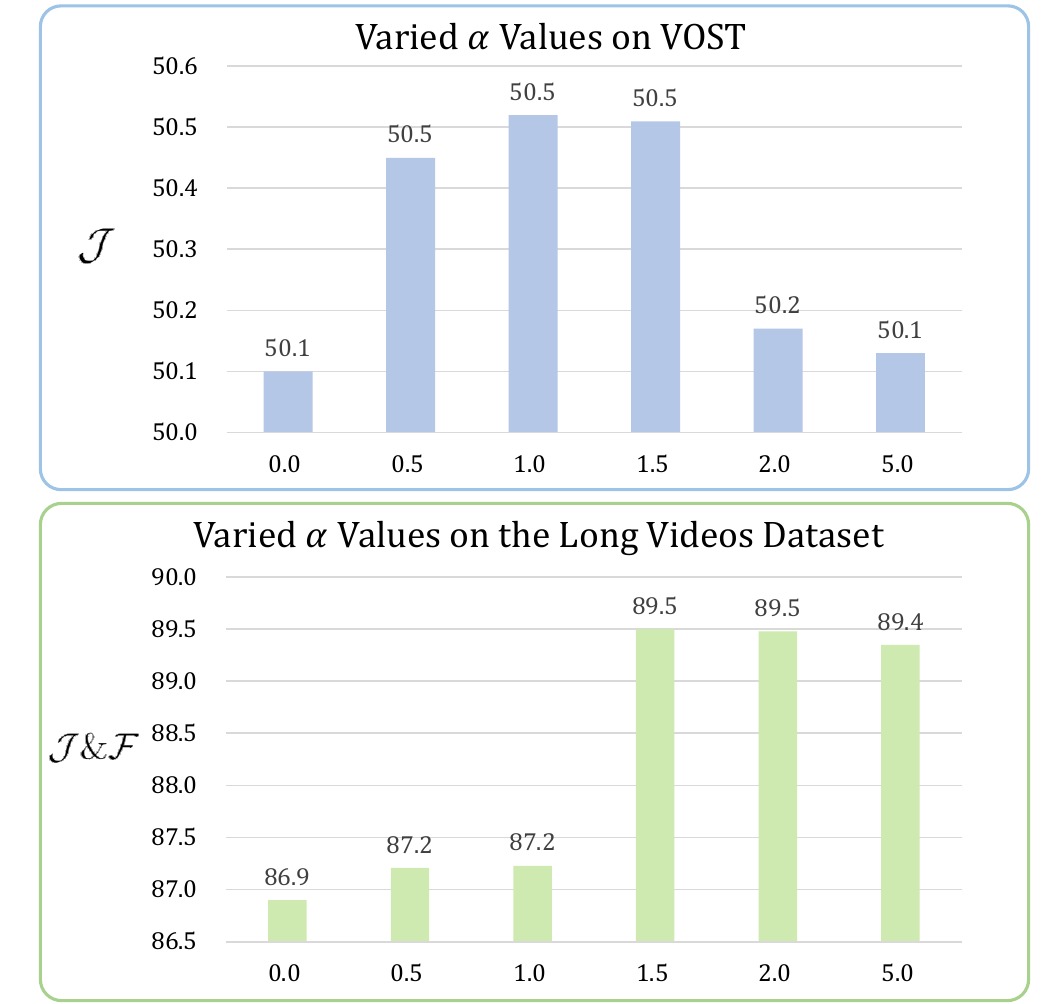}
    \vspace{-2mm}
    \caption{Analysis on relevance and freshness for memory update, on VOST and the Long Videos dataset. The performance varies with different $\alpha$ values (from Eqn.~\ref{eq:balance_rel_fresh}), and it illustrates the importance of the trade-off between relevance and freshness. }
    \label{fig:balance}
    \vspace{-4mm}
\end{figure}

As mentioned in \refsec{sec:mem_update} and Sec.~\ref{sec:supp_detail_mem_update}, we balance relevance and freshness when updating the memory banks via Eqn.~\ref{eq:balance_rel_fresh}. Fig.~\ref{fig:balance} analyzes the performance under different $\alpha$ values on both VOST and the Long Videos dataset. Specifically, a larger $\alpha$ denotes relying more on the freshness term. A proper $\alpha$ is essential for the UCB-inspired algorithm to improve memory update for both VOST and the Long Videos dataset, and we empirically select $\alpha=1.5$ because it generalizes better to both of the datasets. Interestingly, Fig.~\ref{fig:balance} also reveals the difference between VOST and the Long Videos dataset: VOST has more complex scenarios and highlights the utility of relevance, while the long video dataset relies more on freshness due to its extremely long video duration. Nonetheless, our final $\alpha=1.5$ achieves proper balance for both domains. 

\subsection{Relevance Calculation}
\label{sec:supp_ablation_atten_score}

Our relevance term for memory update uses attention scores to reflect the importance of a frame, similar to previous works~\cite{cheng2022xmem, liang2020video}. However, LSTT has three transformer layers and enables two intuitive strategies of relevance calculation: (1) directly using the 0-th layer; and (2) computing the average attention scores of all the transformer layers. Table ~\ref{tab:ablation_atten_score} compares these two strategies on VOST and the Long Videos dataset. We observe that using the 0-th layer for relevance calculation has an advantage in most of the scenarios. We conjecture that the 0-th transformer has the largest fidelity to the features of images and memory banks. Therefore, our \ourmodel empirically selects the 0-th transformer for relevance, as described in Sec.~\ref{sec:supp_detail_mem_update}.

\begin{table}[ht]
  \centering
  \resizebox{0.99\linewidth}{!}{
    \begin{tabular}
      {
      l@{\hspace{3mm}}|
      c@{\hspace{3mm}}c@{\hspace{3mm}}
      |c@{\hspace{3mm}}c@{\hspace{3mm}}c@{\hspace{3mm}}
      }
      \toprule
      \multirow{2}{*}{Methods} & \multicolumn{2}{c|}{VOST} & \multicolumn{3}{c}{Long Video} \\
      \cline{2-3} \cline{4-6}
      & $\jaccdLast$ & $\jaccd$ & $\jaccd \& \contour$ & $\jaccd$ & $\contour$
      \\
      \midrule
      AOT + RMem\ (0$^{th}$) & \textbf{39.8} & \textbf{50.5} & \textbf{90.3} & \textbf{88.5} & \textbf{92.1}
      \\
      AOT + RMem\ (Mean) & 39.6 & 50.3 & 89.8 & 88.2 & 91.5
      \\
      \midrule
      DeAOT + RMem\ (0$^{th}$) & 40.4 & 51.8 & \textbf{91.5} & \textbf{89.8} & \textbf{93.3}
      \\
      DeAOT + RMem\ (Mean) & \textbf{40.6} & \textbf{52.0} & 90.3 & 88.7 & 92.0
      \\
      \bottomrule
    \end{tabular}
  }
  \vspace{-3mm}
  \caption{Analysis on the relevance calculation. ``0-th'' and ``Mean'' denote using the attention scores from the 0-th transformer layer or the average attention scores from all of the three layers.}
  \vspace{-4mm}
  \label{tab:ablation_atten_score}
\end{table}

\subsection{Analysis on Training-Inference alignment}

As discussed in \refsec{sec:mem_temporal}, the purpose of temporal positional embedding is to align the gap between training and inference, as VOS models are trained on short videos but inferencing on unlimited videos. However, it is also valuable to explore whether it is another approach to address this training-inference gap. We compared our  Restricted Memory (RM) with 2 approaches: \textbf{(1) Longer Memory (LM):} train the model with longer video clips so that the model can fit better on a larger memory bank. \textbf{(2) More Steps (MS):} train the model with more steps. As is shown in \reftab{tab:longer_memory_more_step}, LM certainly is effective in mitigating the training-inference gap, but it is still worse than our RM. MS exhibits overfitting with too many training steps, thus not capable of addressing this issue. However, MS can still gain improvement through our RM, proving our method's effectiveness from another perspective.

\begin{table}[ht]
  \centering
  \resizebox{0.9\linewidth}{!}{
    \begin{tabular}
      {l@{\hspace{2mm}}|
        c@{\hspace{2mm}}
        c@{\hspace{2mm}}|
        c@{\hspace{2mm}}
        c@{\hspace{2mm}}|
        c@{\hspace{2mm}}
        c@{\hspace{2mm}}}
      \toprule
      \multirow{2}{*}{Model} & \multirow{2}{*}{Train\_Mem\_Len} & \multirow{2}{*}{Step} & \multicolumn{2}{c|}{URM} & \multicolumn{2}{c}{RM} \\
      \cline{4-5} \cline{6-7}
      & & & $\jaccdLast$ & $\jaccd$ & $\jaccdLast$ & $\jaccd$ \\
      \midrule
      AOT & 4 & 20k & 37.0 & 49.2 & 38.6 & 50.2 \\
      \midrule
      AOT-LM & 6 & 20k & 38.2 & 49.9 & 39.8 & 50.1 \\
      \midrule
      AOT-MS & 4 & 40k & 36.6 & 48.6 & 37.8 & 48.0 \\
      \bottomrule
    \end{tabular}
  }
  \vspace{-3mm}
  \caption{Analysis of 2 approaches to address training-inference gap. ``URM'' for unrstricted memory and ``RM'' for restricted memory. Our ``RM'' is still the best way to align training and inference.}
  \vspace{-4mm}
  \label{tab:longer_memory_more_step}
\end{table}

\subsection{Analysis on LVOS}
\label{sec:result_lvos}

Since the Long Videos dataset only features 3 testing videos, which is not able to fully demonstrate the effectiveness of our method, we further report our model's performance on LVOS dataset \cite{hong2023lvos}, which contains 50 long videos in the validation set.

\begin{table}[ht]
  \centering
  \resizebox{0.7\linewidth}{!}{
    \begin{tabular}
      {
      l@{\hspace{3mm}}|
      c@{\hspace{3mm}}c@{\hspace{3mm}}c@{\hspace{3mm}}
      }
      \toprule
      Methods    & $\jaccd \& \contour$ & $\jaccd$      & $\contour$
      \\
      \midrule
      AOT        & 63.6                 & 57.6          & 69.5
      \\
      AOT + TPE  & 64.5                 & 58.9          & 70.0
      \\
      AOT + RMem & \textbf{66.1}        & \textbf{60.5} & \textbf{71.7}
      \\
      \bottomrule
    \end{tabular}
  }
  \vspace{-3mm}
  \caption{Results on the validation set of LVOS dataset.}
  \vspace{-4mm}
  \label{tab:lvos_result}
\end{table}

As is shown in \reftab{tab:lvos_result}, our RMem still holds the highest performance compared to the AOT baseline. Besides, our TPE (temporal positional embedding) exhibits considerable improvements, which proves that TPE is effective in aligning the training-inference gap, given that the average duration in LVOS is much longer than other video datasets.

\subsection{Analysis on YoutubeVOS2019}

Our study concentrates on improving the VOS accuracy for long and/or complex VOS scenarios. Meanwhile, we also supplement with analysis on shorter, simpler benchmarks. As indicated in \reftab{tab:ablation_short_video}, our \ourmodel demonstrates comparable performance to baselines without \ourmodel on DAVIS2017~\cite{pont20172017}, with a notable increase in efficiency. This result underlines the adaptability of our approach across different regimes.

Further analysis is conducted in the section using the YoutubeVOS2019~\cite{xu2018youtube} benchmark, with shorter video duration and easier scenarios. In Table~\ref{tab:youtube_ablation_deaot}, we evaluate two settings: (1) the influence of only restricting the memory bank sizes; and (2) the effect of the full \ourmodel with temporal positional embedding. Table~\ref{tab:youtube_ablation_deaot} (rows 1 and 2) shows that: by limiting the memory banks with the \emph{original checkpoint} provided by DeAOT's authors, we maintain the same VOS quality. This finding suggests that \emph{constraining the memory banks is a regime-independent strategy}. 

A key aspect of our \ourmodel is temporal positional embedding (TPE), which necessitates end-to-end model training on extended sequences. As in Sec.~\ref{sec:supp_training}, we \emph{increase the training sequence length from 5 frames to 9 frames} without tuning the hyper-parameters, ensuring a 4-frame memory bank during the training stage. However, this introduces optimization challenges, as reflected in the decreased DeAOT performance with longer training clips (Table~\ref{tab:youtube_ablation_deaot}, rows 1 and 3). Under such a setup and fair comparison, our full \ourmodel has maintained comparable VOS quality compared with the baseline (rows 3 and 4). In conclusion, our \ourmodel is also applicable for YoutubeVOS2019, although tuning the optimal hyper-parameters for training with longer sequence lengths is future work.

\begin{table}[ht]
\centering
\resizebox{0.99\linewidth}{!}{
\begin{tabular}
{
c@{\hspace{2mm}}|l@{\hspace{3mm}}|
c@{\hspace{3mm}}c@{\hspace{3mm}}c@{\hspace{3mm}}c@{\hspace{3mm}}c@{\hspace{3mm}}
}
\toprule
Index & Method & $\mathcal{G}$ & $\jaccd_s$ & $\jaccd_u$ & $\contour_s$ & $\contour_u$ \\
\midrule
1 & DeAOT & 85.9 & 84.6 & 89.4 & 80.8 & 88.9 \\
2 & DeAOT + \ourmodel & 85.9 & 84.6 & 89.4 & 80.8 & 88.9 \\
\midrule
3 & DeAOT$^\Psi$ & 85.6 & 84.8 & 80.0 & 89.7 & 88.0 \\
4 & DeAOT$^\Psi$ + \ourmodel & 85.5 & 84.6 & 79.8 & 89.4 & 88.2 \\
\bottomrule
\end{tabular}
}
\vspace{-3mm}
\caption{Analysis on YoutubeVOS2019 shows that, although not the primary focus of this paper, our \ourmodel is also applicable for YoutubeVOS2019 with comparable performance with baselines. We first apply restricted memory banks to the original DeAOT checkpoint (rows 1 and 2). To enable temporal positional embedding (TPE), we train DeAOT under a longer sequence length and denote such models with ``$\Psi$'' (rows 3 and 4). The sub-scripts ``s'' and ``u'' denote the ``seen'' and ``unseen'' subsets of YoutubeVOS2019, respectively.}
\vspace{-5mm}
\label{tab:youtube_ablation_deaot}
\end{table}

\section{Additional Discussion on Limitations and Future Work}
\label{sec:supp_limitations}

We briefly outlined the limitations of our study in \refsec{sec:conclusion} due to space limits. This section elaborates on more details.

As mentioned in \refsec{sec:conclusion}, we prioritize the analysis of memory banks, and \ourmodel is designed as a straightforward instantiation to demonstrate our insight. For this purpose, our study primarily engages with state-of-the-art methods like AOT~\cite{yang2021aot} and DeAOT~\cite{yang2022decoupling}. This choice is grounded, especially when common VOS studies are built upon a single or few preceding approaches due to the complexity of the framework, such as XMem~\cite{cheng2022xmem}, HODOR~\cite{athar2022hodor}, and DeAOT~\cite{yang2022decoupling}. One potential limitation could be that our \ourmodel might implicitly depend on the transformer mechanisms and the affinity calculation in self-attention, which are adopted in AOT and DeAOT. These mechanisms natively support the temporal positional embedding and align with our key motivation of focusing the attention scores on relevant frames (\refsec{sec:pilot} and \reffig{fig:pilot_experiments}). While future endeavors could explore adapting \ourmodel for various VOS methods beyond the ones using transformers, near-future VOS methods will likely continue to employ a transformer-based framework, making our current \ourmodel design compatible with them.

Another aspect mentioned in \refsec{sec:conclusion} is the potential for enhancing \ourmodel with more advanced techniques. While the current simplicity of our approach effectively demonstrates our core insights into managing memory bank capacities, we acknowledge that it can benefit from a more sophisticated design. As especially pointed out in \refsec{sec:conclusion}, XMem~\cite{cheng2022xmem} exhibits an intricate design for efficiently expanding memory banks. Though more complex than our current method of simply bounding memory bank sizes, such advancements could offer greater flexibility and potentially improve VOS.

Lastly, as discussed in \refsec{sec:conclusion}, another option for enhancement lies in improving the decoding capabilities of the VOS framework. Our study maintains the original design of existing methods for a fair comparison, yet future research could explore scaling or modifying VOS architectures to further mitigate the challenges posed by expanding memory banks.


\end{document}